\documentclass[11pt]{article}

\usepackage{authblk}
\usepackage{fullpage}
\usepackage{hyperref}
\usepackage{graphicx}

\title{Cognitive Subscore Trajectory Prediction in Alzheimer's Disease}
\author[1]{Lev E. Givon}
\author[1]{Laura J. Mariano}
\author[1]{David O'Dowd}
\author[1]{John M. Irvine}
\author[1]{Abraham R. Schneider}
\author[ ]{the Alzheimer Disease Neuroimaging Initiative
  \thanks{Data used in preparation of this article were obtained from
    the Alzheimer's Disease Neuroimaging Initiative (ADNI) database
    (\url{http://adni.loni.usc.edu}). As such, the investigators
    within the ADNI contributed to the design and implementation of
    ADNI and/or provided data but did not participate in analysis or
    writing of this report. A complete listing of ADNI investigators
    can be found at:
    \url{http://www.loni.ucla.edu/ADNI/Collaboration/ADNI/Manuscript/Citations.pdf}}}
\affil[1]{The Charles Stark Draper Laboratory, Inc.}
\date{\today}

\begin{document}
\maketitle



\section{Introduction}

Accurate diagnosis of Alzheimer's Disease (AD) entails clinical
evaluation of multiple cognition metrics and biomarkers. Metrics such
as the Alzheimer's Disease Assessment Scale-Cognitive test (ADAS-cog)
\cite{rosen_new_1984} comprise multiple subscores that quantify
different aspects of a patient's cognitive state such as learning,
memory, and language production/comprehension. Although computer-aided
diagnostic techniques for classification of a patient's current
disease state exist \cite{hosseini-asl_alzheimers_2016}, they provide
little insight into the relationship between changes in brain
structure and different aspects of a patient's cognitive state that
occur over time in AD.

\section{Methods}

\begin{figure}
  \centering
  \includegraphics[scale=0.35]{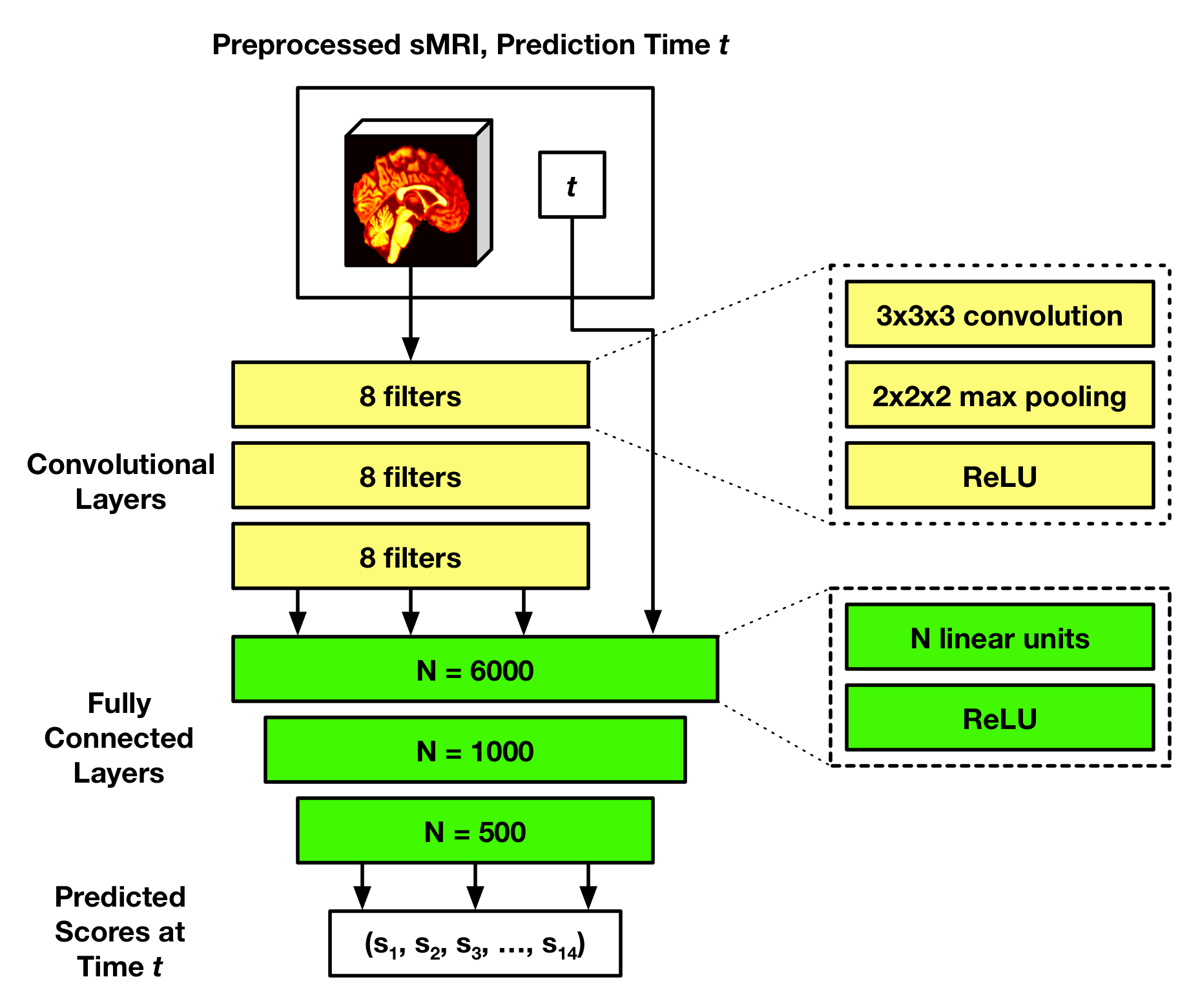}
  \caption{Convolutional neural network architecture for prediction
    of multiple cognitive subscores.}
  \label{fig:cnn}
\end{figure}

We created a Convolutional Neural Network (CNN) architecture that maps
an input tuple comprising a patient's current structural MRI (sMRI)
scan and a future time (in number of months) to the values of the
patient's 13 ADAS-Cog subscores predicted for that time. The
architecture (Fig.~\ref{fig:cnn}) consists of
\begin{itemize}
\item 3 convolutional layers, each comprising convolution, maximum pooling, and
  rectified linear activation;
\item 3 fully connected layers, each comprising $N$ linear units and
  rectified linear activation, where $N=6000,1000,500$, respectively.
\end{itemize}
The first convolutional layer's input consists of the specified sMRI
scan; all outputs of the third convolutional layer and the input
prediction time are fed to all units in the first fully connected
layer. Weights in all convolutional layers were initialized as
described in \cite{he_delving_2015}. To prevent overfitting, dropout
\cite{srivastava_dropout:_2014} with a probability of 0.5 was
performed after max pooling in every convolutional layer during
training. Optimization was performed using the RMSProp
\cite{hinton_introduction_2014} algorithm. A smooth $L_1$ loss
function was employed to reduce sensitivity to outliers
\cite{girshick_fast_2015}. The architecture was implemented in
PyTorch \cite{paszke_pytorch_2017} and trained on Ubuntu Linux 16.04
with NVIDIA Tesla M40 Graphics Processing Units (GPUs).

To evaluate the architecture, we performed 5-fold cross-validation
with training and testing data drawn from a set of tuples of sMRI
scans, cognitive scores, and time intervals between MRI and cognitive
score acquisition. Intervals were multiples of 6 months between 0 and
36. Each fold was stratified to contain equal numbers of subscores for
each unique interval. To ensure that each fold contained a sufficient
number of tuples for each interval, the folds were drawn from a
dataset comprising 1000 entries. This dataset contained multiple
images for some patients, although the number of images obtained from
an individual patient across the dataset was not fixed owing to the
size of the parent ADNI dataset from which the images were drawn.

All sMRI and cognitive score data used for evaluation of the
architecture were obtained from the Alzheimer’s Disease Neuroimaging
Initiative (ADNI) database (\url{http://adni.loni.usc.edu}). The
primary goal of ADNI has been to test whether serial MRI, PET, other
biological markers, and clinical and neuropsychological assessment can
be combined to measure the progression of mild cognitive impairment
(MCI) and early Alzheimer’s disease (AD). Only 1.5T MRI scans from a
quality-controlled subset of the ADNI1 phase of the project were
utilized \cite{wyman_standardization_2013}. sMRI scans were minimally
preprocessed to remove non-brain tissue, normalize intensities, and
register all scans to a single coordinate space. All subscores were
normalized to the range $[0,1]$ prior to training and testing.

\section{Results}

\begin{figure}
  \centering
  \includegraphics[scale=0.35]{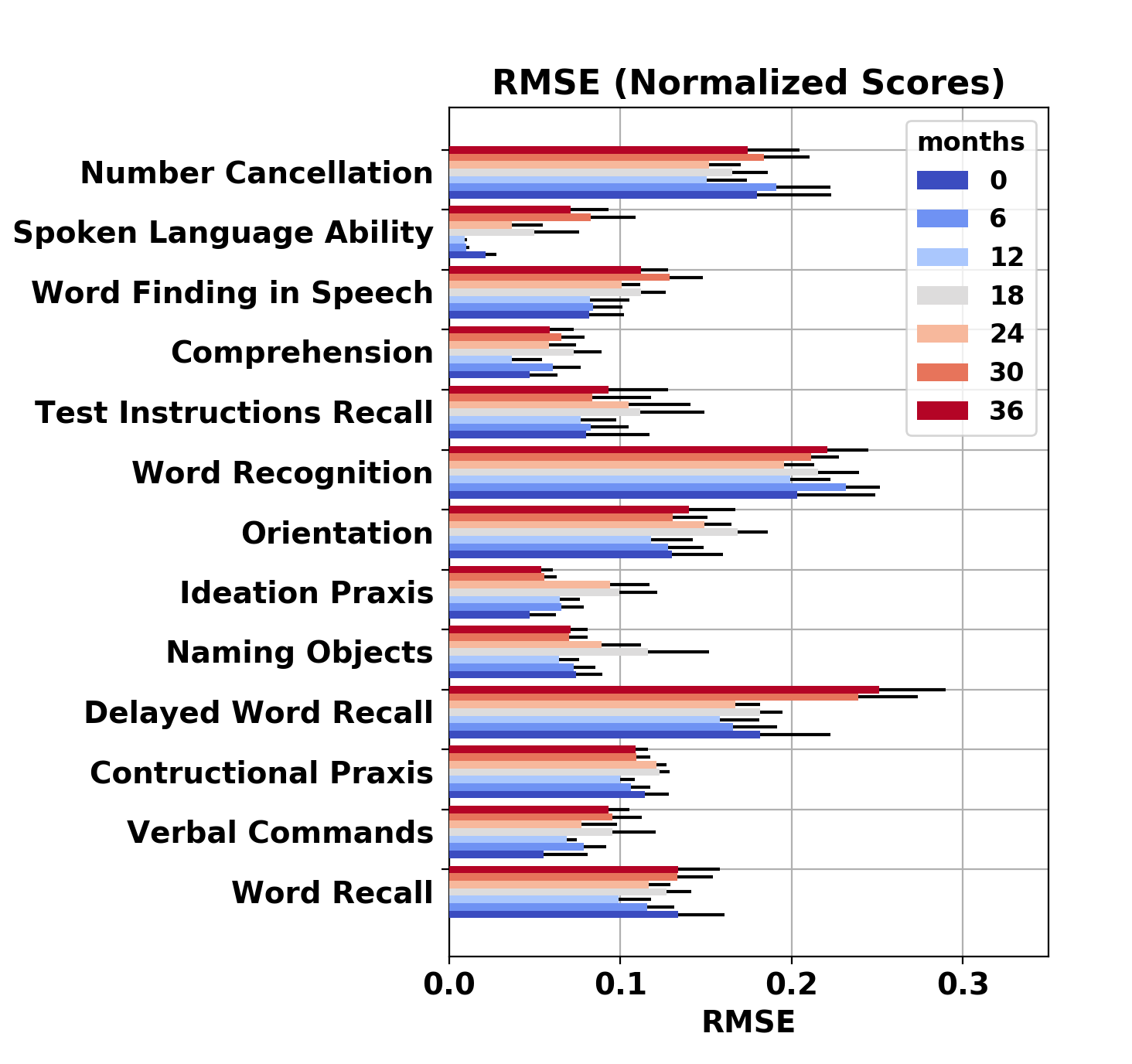}
  \caption{RMSEs of predicted normalized subscores.}
  \label{fig:rmse}
\end{figure}

\begin{figure}
  \centering
  \includegraphics[scale=0.35]{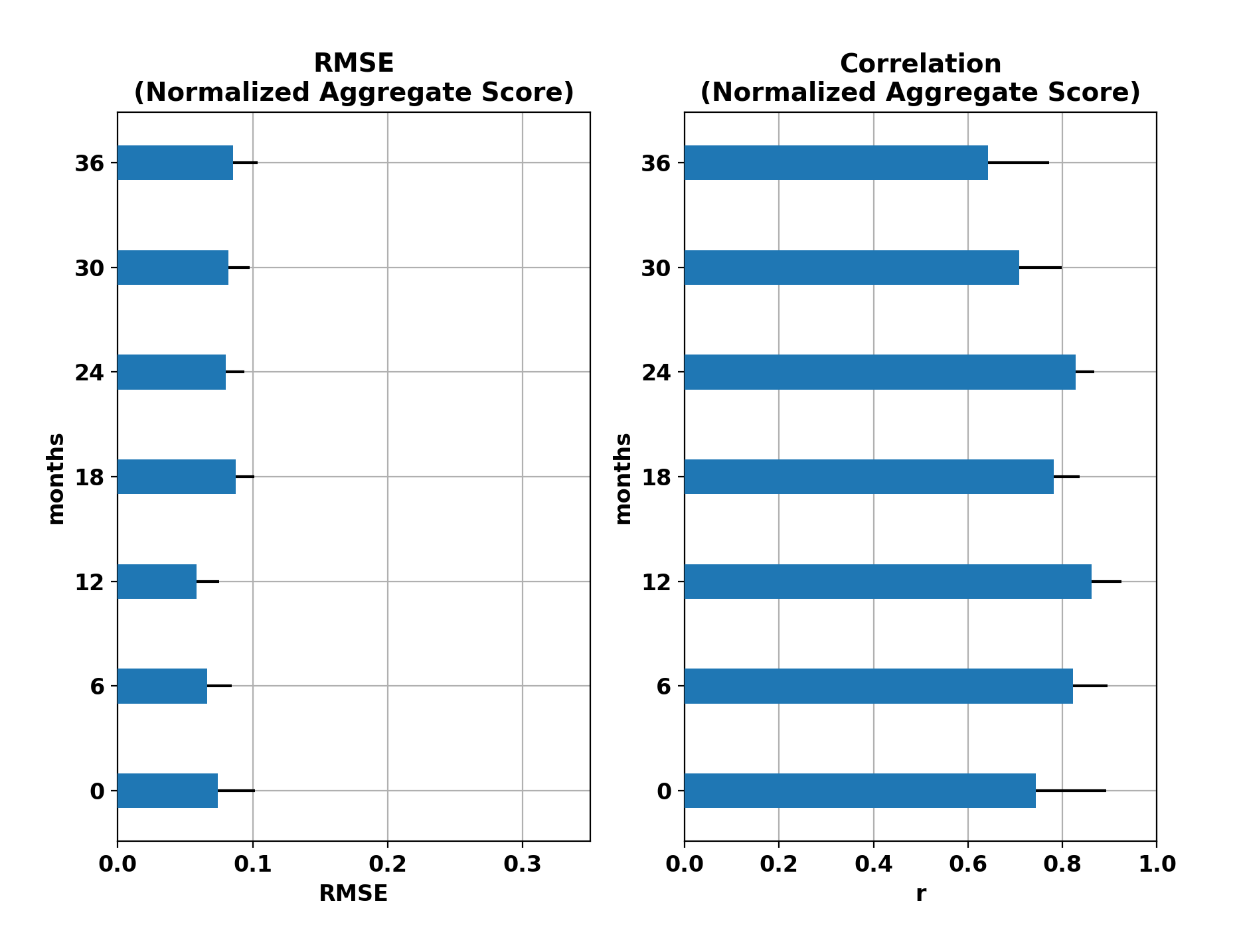}
  \caption{RMSEs and Pearson's correlation coefficients of predicted
    normalized aggregate scores.}
  \label{fig:agg}
\end{figure}

We computed the root mean squared error (RMSE) for each of the 13
subscores for each interval (Fig.~\ref{fig:rmse}). The means and
standard errors across all intervals and subscores varied from 0.009
to 0.251 and from 0.001 to 0.046, respectively. To compare the
performance of our network with other approaches that only predict the
aggregate score, we also computed the RMSE and Pearson's correlation
coefficient of the predicted/actual aggregate score normalized to the
range $[0,1]$ (Fig.~\ref{fig:agg}). We obtained the latter by
rescaling the normalized subscores to their original respective ranges
and normalizing the sum using the highest possible aggregate
score. The means and standard errors of the RMSEs of our method's
normalized aggregate score prediction for the intervals considered
varied from 0.058 to 0.087 and from 0.014 to 0.028, respectively. The
means and standard errors of the correlation between predicted and
actual aggregate scores across all intervals varied from 0.643 to
0.862 and from 0.040 to 0.149, respectively.

\section{Discussion}

To our knowledge, our architecture is the first to concurrently
predict multiple cognitive examination subscores from minimally
preprocessed structural brain data. Variations in prediction accuracy
across the subscores illustrate differences in the relationships
between brain structure and specific aspects of cognition that are
obscured by prediction of the aggregate score. The mean performance of
our architecture when applied to aggregate score prediction is similar
to that of existing techniques that variously utilize stepwise
regression \cite{walhovd_multi-modal_2010}, relevance vector
regression \cite{stonnington_predicting_2010}, multi-task learning
\cite{wang_sparse_2011, wan_sparse_2012, zhou_modeling_2013}, and
support vector machines \cite{zhang_multi-modal_2012}. Since the
quality of features extracted by CNNs is proportional to the amount of
training data and number of network layers
\cite{krizhevsky_imagenet_2012}, we anticipate that deeper variations
of our architecture trained on more extensive subsets of the ADNI
database will be able to achieve state-of-the-art performance.

\section{Conclusion}

We have developed a CNN architecture that can predict the trajectories
of the 13 subscores comprised by a subject's ADAS-cog examination
results from a current sMRI scan up to 36 months from image
acquisition time without resorting to manual feature extraction. Mean
performance metrics are within range of those of existing techniques
that require manual feature selection and are limited to predicting
aggregate scores.

\section{Keywords}

Alzheimer's Disease, deep learning, convolutional neural network,
cognitive assessment, machine learning, MRI

\section{Acknowledgments}

Data collection and sharing for this project was funded by the
Alzheimer's Disease Neuroimaging Initiative (ADNI, principal
investigator: Michael Weiner) (National Institutes of Health Grant U01
AG024904) and DOD ADNI (Department of Defense award number
W81XWH-12-2-0012). ADNI is funded by the National Institute on Aging,
and the National Institute of Biomedical Imaging and Bioengineering.
ADNI data are disseminated by the Laboratory for Neuro Imaging at the
University of Southern California.

\bibliography{references}
\bibliographystyle{plain}

\end{document}